\documentclass[10pt,twocolumn,letterpaper]{article}
\usepackage{cvpr}
\usepackage{times}
\usepackage{epsfig}
\usepackage{graphicx}
\usepackage{amsmath}
\usepackage{amssymb}
\usepackage{color}
\usepackage{multirow}
\usepackage{makecell}
\usepackage{subcaption}
\usepackage[noadjust]{cite}

\newlength{\blackoutwidth}

\newcommand{\Lagr}{\mathcal{L}}

\DeclareMathOperator*{\argmin}{arg\,min\,\,max}

\usepackage[pagebackref=true,breaklinks=true,letterpaper=true,colorlinks,bookmarks=false]{hyperref}
\usepackage{romannum}

\usepackage{algorithm}
\usepackage[noend]{algpseudocode}
\usepackage{amsmath}
\makeatletter
\def\BState{\State\hskip-\ALG@thistlm}
\makeatother

\cvprfinalcopy 

\def\cvprPaperID{68} 

\ifcvprfinal\fi
\begin{document}
\pagenumbering{arabic}
\pagenumbering{gobble}

\title{Cycle-Dehaze: Enhanced CycleGAN for Single Image Dehazing} 

\author{Deniz Engin\thanks{indicates equal contribution} \qquad \qquad An{\i}l Gen\c{c}\footnotemark[1] \qquad \qquad Haz{\i}m Kemal Ekenel\\
SiMiT Lab, Istanbul Technical University, Turkey\\
{\tt\small \{deniz.engin, genca16, ekenel\}@itu.edu.tr}
}

\maketitle

\begin{abstract}
In this paper, we present an end-to-end network, called Cycle-Dehaze, for single image dehazing problem, which does not require pairs of hazy and corresponding ground truth images for training. That is, we train the network by feeding clean and hazy images in an unpaired manner. Moreover, the proposed approach does not rely on estimation of the atmospheric scattering model parameters. Our method enhances CycleGAN formulation by combining cycle-consistency and perceptual losses in order to improve the quality of textural information recovery and generate visually better haze-free images. Typically, deep learning models for dehazing take low resolution images as input and produce low resolution outputs. However, in the NTIRE 2018 challenge on single image dehazing, high resolution images were provided. Therefore, we apply bicubic downscaling. After obtaining low-resolution outputs from the network, we utilize the Laplacian pyramid to upscale the output images to the original resolution. We conduct experiments on NYU-Depth, I-HAZE, and O-HAZE datasets. Extensive experiments demonstrate that the proposed approach improves CycleGAN method both quantitatively and qualitatively.
\end{abstract}
\section{Introduction}

\begin{figure}[t!]
 \centering
     \begin{subfigure}[b]{0.23\textwidth}
             \includegraphics[width=\textwidth,height=27mm]{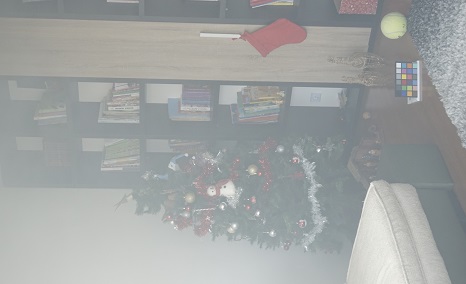}
     \end{subfigure}
     \begin{subfigure}[b]{0.23\textwidth}
             \includegraphics[width=\textwidth,height=27mm]{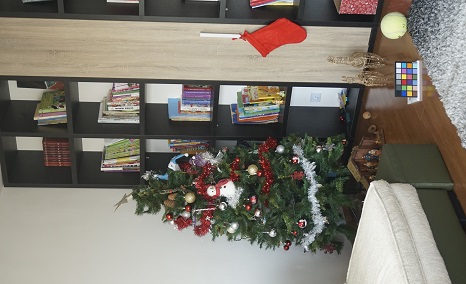}
     \end{subfigure}
     \\
     \vspace{0.5mm}
     \begin{subfigure}[b]{0.23\textwidth}            
             \includegraphics[width=\textwidth,height=36mm]{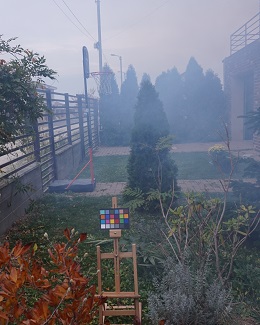}
     \end{subfigure}
      \begin{subfigure}[b]{0.23\textwidth}
             \includegraphics[width=\textwidth,height=36mm]{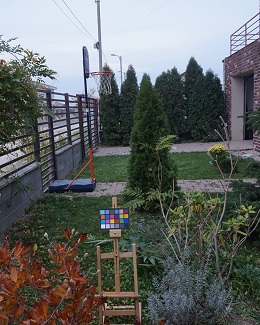}
     \end{subfigure}
     \caption{Hazy and clean examples from the NTIRE 2018 challenge on single image dehazing datasets: I-HAZE~\cite{indoorNTIREdataset} \& O-HAZE~\cite{outdoorNTIREdataset} datasets.}
     \label{example-hazy}
 \end{figure}

Bad weather events such as fog, mist, and haze dramatically reduce the visibility of any scenery and constitute significant obstacles for computer vision applications, \eg object detection, tracking, and segmentation. While images captured from hazy fields usually preserve most of their major context, they require some visibility enhancement as a pre-processing before feeding them into computer vision algorithms, which are mainly trained on the images captured at clear weather conditions. This pre-processing is generally called as \emph{image dehazing/defogging}. Image dehazing techniques aim to generate haze-free images purified from the bad weather events. Sample hazy and haze-free images from the NTIRE 2018 challenge on single image dehazing~\cite{ChallengeReportAncuti_2018_CVPR_Workshops} are illustrated in Figure \ref{example-hazy}.

In the recent literature, researchers concentrate on single image dehazing methods, which can dehaze an input image without requiring any extra information, \eg depth information or known 3D model of the scene. Single image dehazing approaches are divided into prior information-based methods~\cite{new1tarel2009fast,new2ancuti2010fast,dcp2011,new5meng2013efficient,new3emberton2015hierarchical,new4ancuti2016night} and learning based methods~\cite{cai2016dehazenet,reneccv2016singleCNN,zhang2017joint,yang2018towards,swami2018candy}. Prior information-based methods are mainly based on the parameter estimation of atmospheric scattering model by utilizing the priors, such as dark channel priors~\cite{dcp2011}, color attenuation prior~\cite{CAP2015fast}, haze-line prior~\cite{berman2016non,berman2017air}. On the other hand, these parameters are obtained from training data by learning based methods, which rely mostly on deep learning approaches. The proliferation of deep neural networks increases the use of large-scale datasets, therefore, researchers tend to create synthetic dehazing datasets like \emph{FRIDA~\cite{FRIDA} and D-HAZY~\cite{Dhazydata2016}}, which have a more practical creation process than real dehazing datasets. Even though most of the deep learning approaches use the estimation of intermediate parameters, \eg transmission map and atmospheric light~\cite{cai2016dehazenet,reneccv2016singleCNN}, there are also other approaches based on generative adversarial networks (GANs), which build a model without benefiting from these intermediate parameters~\cite{swami2018candy}.

GANs, introduced by Goodfellow \etal~\cite{goodfellow2014generative}, are found to be very successful at image generation tasks, \eg data augmentation, image inpainting, and style transfer. Their major goal is the generation of fake images indistinguishable from the original images on the targeted domain. By utilizing GANs, there exist \emph{state-of-the-art} methods~\cite{swami2018candy,zhang2017joint} for single image dehazing, which require hazy input image and its ground truth in a paired manner. Recently, the need of paired data is removed after the cycle-consistency loss has been proposed by CycleGAN~\cite{cyclegan} for image-to-image translation. Inspired by the cycle-consistency loss, Disentangled Dehazing Network (DDN) has been introduced by Yang \etal for single image dehazing. Unlike CycleGAN~\cite{cyclegan} architecture, DDN reconstructs cyclic-image via the atmospheric scattering model instead of using another generator. Therefore, it requires the scene radiance, medium transmission map, and global atmospheric light~\cite{yang2018towards} at the training phase. 

In this work, we introduce Cycle-Dehaze network by utilizing CycleGAN~\cite{cyclegan} architecture via aggregating cycle-consistency and perceptual losses. Our main purpose is building an end-to-end network regardless of atmospheric scattering model for single image dehazing. In order to feed the input images into our network, they are resized to $256 \times 256$ pixel resolution via bicubic downscaling. After dehazing the input images, bicubic upscaling to their original size is not sufficient to estimate the missing information. To be able to obtain high-resolution images, we employed a simple upsampling method based on Laplacian pyramid. We perform our experiments on NYU-Depth~\cite{NYUdataset} part of D-HAZY~\cite{Dhazydata2016} and the NTIRE 2018 challenge on single image dehazing~\cite{ChallengeReportAncuti_2018_CVPR_Workshops} datasets: I-HAZE~\cite{indoorNTIREdataset} \& O-HAZE~\cite{outdoorNTIREdataset}. According to our results, Cycle-Dehaze achieves higher image quality metrics than CycleGAN~\cite{cyclegan} architecture. Moreover, we analyze the performance of Cycle-Dehaze on cross-dataset scenarios, that is, we use different datasets at training and testing phases. \\

Our main contributions are summarized as follows:

\begin{itemize}  
\item We enhance CycleGAN \cite{cyclegan} architecture for single image dehazing via adding cyclic perceptual-consistency loss besides cycle-consistency loss.
\item Our method requires neither paired samples of hazy and ground truth images nor any parameters of atmospheric scattering model during the training and testing phases. 
\item We present a simple and efficient technique to upscale dehazed images by benefiting from Laplacian pyramid.
\item Due to its cyclic structure, our method provides a generalizable model demonstrated with the experiments on cross-dataset scenarios.
\end{itemize}

The rest of this paper is organized as follows: In Section~\ref{related}, a brief overview of related work is provided. The proposed method is described in Section~\ref{method}. Experimental results are presented and discussed in Section~\ref{experiment}. Finally, the conclusions are given in Section~\ref{conc}.
\section{Related Work} \label{related}

Image dehazing methods aim at recovering the clear scene reflection, atmospheric light color, and transmission map from an input hazy image. The requirement to know several parameters of the scene makes this problem challenging. Image dehazing methods can be categorized in terms of their inputs: (\romannum{1}) multiple images dehazing, (\romannum{2}) polarizing filter-based dehazing, (\romannum{3}) single image dehazing via utilizing additional information, \eg depth or geometrical information methods, and (\romannum{4}) single image dehazing~\cite{li2017hazesurvey}. 

Multiple images based methods overcome dehazing problem by obtaining changed atmospheric conditions from multiple images~\cite{multiple1nayar1999vision,multiple2imagenarasimhan2000chromatic,multiple3narasimhan2003contrast,multiple4zhang2011atmospheric}. In other words, it is required to wait until the weather condition or haze level are changed; thus, it is not practical for real-world applications. The polarization filter based approach has been proposed to dismiss the requirement of changed weather conditions~\cite{schechner2001instant}. In this approach, various filters are applied on different images to simulate changed weather conditions. Nevertheless, the static scenes are only considered when polarization filter based approaches are used. Therefore, this method still is not applicable for real-time scenarios. To address the necessities of these methods, single image dehazing via using additional information such as depth information~\cite{hautiere2007towards} and the approximation of the 3D model of the scene ~\cite{kopf2008deep} have been suggested. Since there is usually a single captured image of hazy scenes in the real-world conditions, obtaining additional information about the scene is extremely hard. Due to problems of previous approaches, researchers focus on single image dehazing methods.\\

\textbf{Single Image Dehazing.} Single image dehazing methods are mainly based on estimating parameters of the physical model, which is also known as the atmospheric scattering model. This model depends on the atmospheric condition of the scene, and can be expressed as follow:
\begin{equation}
\label{eq1}
I(x) = J(x)t(x) + A(1-t(x)) 
\end{equation}
where $I(x)$ is the hazy image, $J(x)$ is the haze-free image or the scene radiance, $t(x)$ is the medium transmission map, and A is the global atmospheric light on each $x$ pixel coordinates. $t(x)$ can be formulated as:
\begin{equation}
t(x) = e^{-\beta d(x)}
\end{equation}
where $d(x)$ is the depth of the scene point and $\beta$ is defined as the scattering coefficient of the atmosphere.

Single image based methods can be categorized into two main approaches: prior information-based methods and learning-based methods. Prior information-based methods have been introduced as the pioneer of single image dehazing methods in~\cite{tan2008visibility,fattal2008single}. Following these studies, the dark channel prior (DCP) based on the statistics about the haze-free images has been presented by He \etal~\cite{dcp2011}. In this method, haze transmission map is estimated by utilizing dark pixels, which have a low-intensity value of at least one color channel. Dark channel prior has been enhanced by optimizing the inherent boundary constraint with weighted L1-norm contextual regularization to estimate transmission map~\cite{new5meng2013efficient}. In addition, Zhu \etal proposed a color attenuation prior (CAP) in order to recover depth information by creating a linear model on local priors~\cite{CAP2015fast}. Contrary to using local priors, Berman \etal introduced non-local color prior (NCP), which is based on an approximation of the \textit{entire} haze-free images including few hundred distinct colors~\cite{berman2016non}. Each distinct color on a haze-free image is clustered and represented a line in RGB color space. Distance map and dehazed image are obtained by using these lines. Haze-line prior based approach has been improved by intersection with air-light to estimate global air-light in~\cite{berman2017air}. Moreover, due to non-uniform lighting conditions on the entire image, local airlight is estimated for each patch for night-time dehazing by multi-scale fusion in~\cite{new4ancuti2016night}.


Recently, learning based methods have been employed by utilizing CNNs and GANs for single image dehazing. CNN based methods mainly focus on estimating transmission map and/or atmospheric light~\cite{cai2016dehazenet,berman2016non} to recover clean images via atmospheric scattering model. On the other hand, GANs based methods produce haze-free images and estimate parameters of the physical model~\cite{yang2018towards}. Also, combination of them has been proposed in~\cite{zhang2017joint}.

Ren \etal~\cite{reneccv2016singleCNN} proposed a Multi-Scale CNN (MSCNN), which consists of coarse-scale and fine-scale networks in order to estimate the transmission map. The coarse-scale network estimates the transmission map, which is also improved locally by the fine-scale network. Another transmission map estimation network, called as DehazeNet, is designed differently from classical CNNs by adding feature extraction and non-linear regression layers, and it has been suggested by Cai \etal~\cite{cai2016dehazenet}. In addition to previous approaches, All-In-One Dehazing Network (AOD-net) has been presented in~\cite{li2017aodneticcv} to be able to produce clean images directly without estimating intermediate parameters independently. Atmospheric scattering model is re-formulated to implement it in an end-to-end network. Zhang \etal~\cite{zhang2017joint} suggested a multi-task method that includes three modules, which are transmission map estimation via GANs, hazy feature extraction, and image dehazing. All modules have been trained jointly and image level loss functions, \eg perceptual loss and pixel-wise Euclidean loss, have been utilized. Similarly, Yang \etal~\cite{yang2018towards} introduced Disentangled Dehazing Network (DDN) to estimate the scene radiance, transmission map, and global atmosphere light by utilizing three generators jointly. Different from our method, these methods require estimation parameters of the atmospheric scattering model during training phase.


\section{Proposed Method} \label{method}

\begin{figure*}[t!]
\begin{center}
\includegraphics[width=0.93\linewidth,keepaspectratio=true]{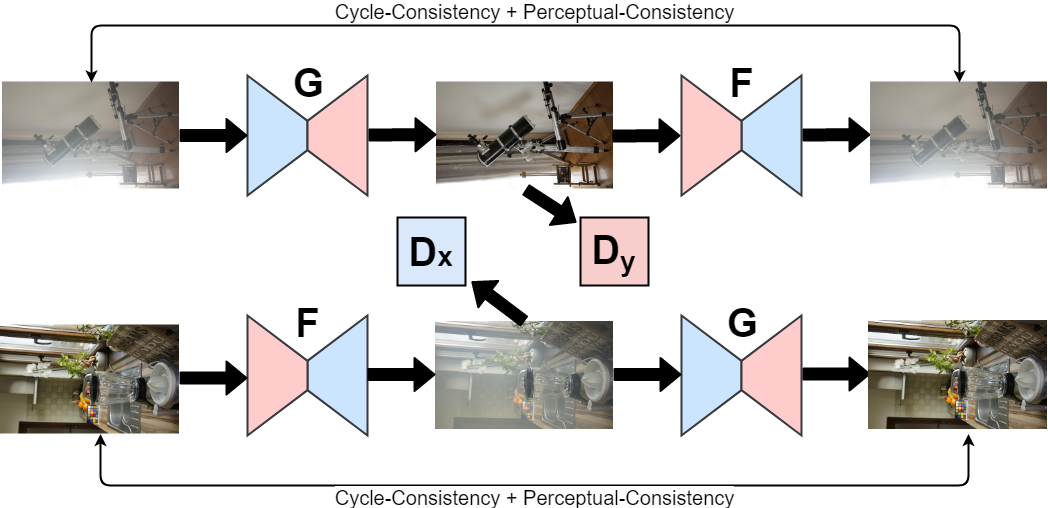}
\end{center}
   \caption{The architecture of Cycle-Dehaze Network where $G$ \& $F$ refers to the generators, and $D_x$ \& $D_y$ to the discriminators. For the sake of clarity, the representation is split into two parts: hazy to clean image, and clean to hazy image. {\color{red} Best view in color.} }
\label{fig:method}
\end{figure*}

\emph{Cycle-Dehaze} is an enhanced version of CycleGAN~\cite{cyclegan} architecture for single image dehazing. In order to increase visual quality metrics, \textit{PSNR, SSIM}, it utilizes the perceptual loss inspired by EnhanceNet~\cite{enhancenet}. The main idea of this loss is comparing images in a feature space rather than in a pixel space. Therefore, Cycle-Dehaze compares the original image with the reconstructed cyclic-image at both spaces, where cycle-consistency loss ensures a high PSNR value and perceptual loss preserves the sharpness of the image. Moreover, Cycle-Dehaze uses traditional Laplacian pyramid to provide better upsampling results after the main dehazing process. Figure \ref{fig:method} shows the overall representation of Cycle-Dehaze architecture. 

As can be demonstrated in Figure \ref{fig:method}, Cycle-Dehaze architecture is composed of two generators \textit{$G,F$} and two discriminators \textit{$D_x,D_y$}. In favor of cleaning/adding the haze, the architecture profits from the combination of cycle-consistency and cyclic perceptual-consistency losses besides the regular discriminator and generator losses. As a result of this, the architecture is forced to preserve textural information of the input images and generate unique haze-free outputs. On the other hand, pursuing the balance between cycle-consistency and perceptual-consistency losses is not a trivial task. Giving over-weight to perceptual loss causes the loss of color information after dehazing process. Therefore, cycle-consistency loss needs to have higher weights than the perceptual loss.\\

\textbf{Cyclic perceptual-consistency loss.} CycleGAN~\cite{cyclegan} architecture introduces cycle-consistency loss, which calculates $L1-norm$ between the original and cyclic image for unpaired image-to-image translation task. However, this calculated loss between the original and cyclic image is not enough to recover all textural information, since hazy images are mostly heavily-corrupted.  Cyclic perceptual-consistency loss aims to preserve original image structure by looking the combination of high and low-level features extracted from $2nd$ and $5th$ pooling layers of \emph{VGG16}~\cite{vgg} architecture. Under the constraints of $x \in X$, $y\in Y$ and generator $G:X\rightarrow Y$, generator $F:Y\rightarrow X$, the formulation of cyclic perceptual-consistency loss is given below, where ($x$, $y$) refers to hazy and ground truth unpaired image set and $\phi$ is a \emph{VGG16}~\cite{vgg} feature extractor from $2nd$ and $5th$ pooling layers: 
\begin{equation}
\label{eq2}
\begin{split}
\Lagr_{Perceptual} &= \Arrowvert \phi(x)-\phi(F(G(x))) \Arrowvert_2^2 \\ 
&+ \Arrowvert \phi(y)-\phi(G(F(y))) \Arrowvert_2^2. 
\end{split}
\end{equation}

\textbf{Full objective of Cycle-Dehaze.} Cycle-Dehaze has one extra loss compared to CycleGAN \cite{cyclegan} architecture. Therefore, the objective of Cycle-Dehaze can be formulated as follows, where $\Lagr_{CycleGAN}(G,F,D_x,D_y)$ is the full objective of CycleGAN~\cite{cyclegan} architecture, $D$ stands for the discriminator and $\gamma$ controls the effect of cyclic perceptual-consistency loss:
\begin{equation}
\label{eq3}
\begin{split}
\Lagr(G,F,D_x,D_y) &= \Lagr_{CycleGAN}(G,F,D_x,D_y) \\
&+ \gamma*\Lagr_{Perceptual}(G, F),
\end{split}
\end{equation} 

\begin{equation}
\label{eq4}
G^*, F^* = \argmin_{G,F,D_x,D_y}\,\Lagr(G,F,D_x,D_y).  
\end{equation} 
Conclusively, Cycle-Dehaze optimizes CycleGAN~\cite{cyclegan} architecture with the additional cyclic perceptual-consistency loss given in Equation~\ref{eq2} according to Equations~\ref{eq3} and~\ref{eq4}. In order to obtain haze-free images, the generator $G^*$ is used at the testing time. 

\textbf{Laplacian upscaling.} Cycle-Dehaze architecture takes $256 \times 256$ pixel resolution input image and produces $256 \times 256$ pixel resolution output image because of GPU limitation. In order to reduce deterioration of the image quality during the downscaling and upscaling process, we have taken advantage of Laplacian pyramid, which is created by using high-resolution hazy images. In order to get the high-resolution dehazed image, we have changed the top layer of Laplacian pyramid with our dehazed low-resolution image and performed Laplacian upscaling process as usual. This basic usage of Laplacian pyramid especially preserves most of the edges of the hazy image during the cleaning process and boosts SSIM value at the upscaling stage. Laplacian upscaling is an optional post-processing step while working on the high-resolution images. \\

\begin{table*}[t!]
  \begin{center}
  \resizebox{\textwidth}{!}{
  \begin{tabular}{c|ccccccccc}
   \hline
   Metrics & None & He \etal~\cite{dcp2011} & Zhu \etal~\cite{CAP2015fast} & Berman \etal~\cite{berman2016non} & Ren \etal~\cite{reneccv2016singleCNN} & Cai \etal~\cite{cai2016dehazenet} & CycleGAN & Yang \etal~\cite{yang2018towards} & \textbf{Ours} \\
  \hline\hline
  PSNR & 9.46 & 10.98 & 12.78 & 12.26 & 13.04 & 12.84 & 13.38 & {\color{red}{15.54}} & {\color{blue}{15.41}}\\
  SSIM & 0.58 & 0.64 & 0.70 & 0.70 & 0.66 & {\color{blue}{0.71}} & 0.52 & {\color{red}{0.77}} & 0.66\\
  \hline
   \end{tabular}}
   \end{center}
   \caption{Average PSNR and SSIM results on NYU-Depth~\cite{NYUdataset} dataset. Most of the accuracies taken from the paper~\cite{yang2018towards}. Numbers in {\color{red}{red}} and {\color{blue}{blue}} indicate first and second best results, respectively. The second column of the table shows the values which are average PSNR and SSIM results calculated directly between the each hazy and its ground truth image.} 
  \label{table:d-hazy}
 \end{table*}
\textbf{Implementation details (indoor/outdoor).} We used TensorFlow~\cite{tensorflow2015-whitepaper} framework for the training and testing phases, and MATLAB for resizing images. We trained our model with NVIDIA TITAN X graphics card. We performed around $40$ epochs on each dataset in order to ensure convergence. Our testing time is about $8$ seconds per image on Intel Core i7-5820K CPU. During the training phase of our model, we used Adam optimizer with the learning rate $1e-4$. Moreover, we took $\gamma$ as $0.0001$ which is $1e+5$ times lower than the weight of the cycle-consistency loss. 

Our network is similar to the original CycleGAN~\cite{cyclegan} architecture except for the cyclic perceptual-consistency loss and Laplacian pyramid as a post-processing step. At the highest level of Laplacian pyramid, we scale the images to $256 \times 256$ pixel resolution because of our network's requirements. To calculate cyclic perceptual-consistency loss, we used VGG16~\cite{vgg} architecture, which is initialized by ImageNet~\cite{imagenet} pre-trained model. The source code of the proposed method will be publicly available through project's GitHub page\footnote{github.com/engindeniz/Cycle-Dehaze}.

\section{Experiments and Results} \label{experiment}


In this section, we have presented the experimental results and discussed them with the results of the challenge. The first experiment is to compare our result with the \emph{state-of-the-art} approaches on NYU-Depth~\cite{NYUdataset} dataset. Then, we have investigated our performance on the NTIRE 2018 challenge on single image dehazing~\cite{ChallengeReportAncuti_2018_CVPR_Workshops} datasets: I-HAZE~\cite{indoorNTIREdataset} \& O-HAZE~\cite{outdoorNTIREdataset}. In addition, we have emphasized differences between CycleGAN~\cite{cyclegan} and our proposed method, \emph{Cycle-Dehaze}, via qualitative and quantitative results. Furthermore, we have provided comparative qualitative results on natural images.

\subsection{Datasets} 

\emph{NYU-Depth}~\cite{NYUdataset} dataset consists of $1449$ pairs of clean and synthesized hazy images of the same scene. The dataset is the part of D-HAZY~\cite{Dhazydata2016} dataset, which includes two individual environments presented as Middelbury~\cite{Middelburydata2014} and NYU-Depth~\cite{NYUdataset}. We have chosen NYU-Depth~\cite{NYUdataset}, which is considerably larger scale than Middlebury~\cite{Middelburydata2014} part. NYU-Depth~\cite{NYUdataset} contains also depth map of each scene, which is not used for this study.

\emph{The NTIRE 2018 challenge on single image dehazing~\cite{ChallengeReportAncuti_2018_CVPR_Workshops}} datasets were collected via professional fog generators and camera setup for image dehazing problem. Each image includes a Macbeth ColorChecker mostly used for color calibration with the real-world. The challenge~\cite{ChallengeReportAncuti_2018_CVPR_Workshops} has two main datasets: I-HAZE~\cite{indoorNTIREdataset} \& O-HAZE~\cite{outdoorNTIREdataset}, which have $25$ indoor and $35$ outdoor hazy images and their ground truth images, respectively. The captured images are in very high resolution. During the challenge, the organizers provide additional $10$ images for each dataset as a validation and test set. We did not include them in the training data. \\


\begin{figure*}[t!]
\centering
    \begin{subfigure}[b]{0.24\textwidth}            
            \includegraphics[width=\textwidth,keepaspectratio=true]{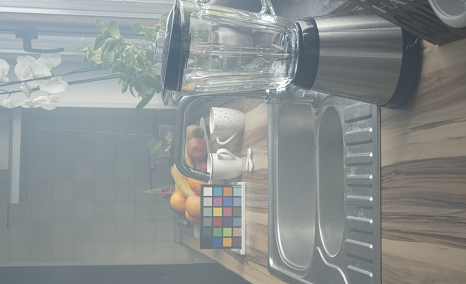}
    \end{subfigure}
     \hspace{0.1mm}
     \begin{subfigure}[b]{0.24\textwidth}
            \includegraphics[width=\textwidth,keepaspectratio=true]{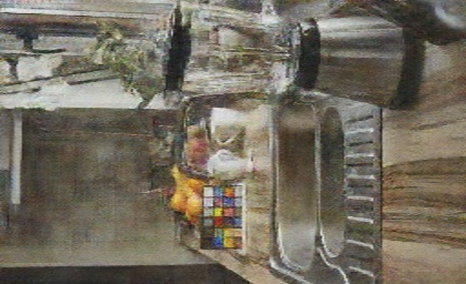}
    \end{subfigure}
     \hspace{0.1mm}
     \begin{subfigure}[b]{0.24\textwidth}
            \includegraphics[width=\textwidth,keepaspectratio=true]{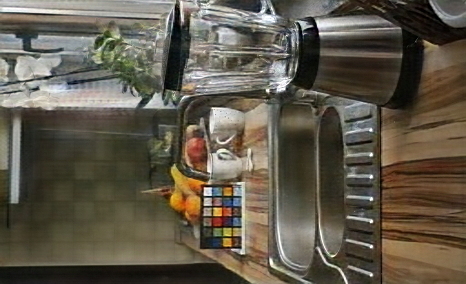}
    \end{subfigure}
   \hspace{0.1mm}
    \begin{subfigure}[b]{0.24\textwidth}
            \includegraphics[width=\textwidth,keepaspectratio=true]{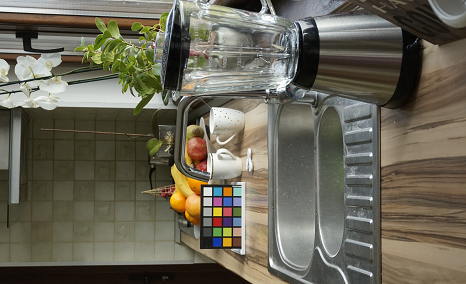}
    \end{subfigure}
    \\
    \vspace{1mm}
    \begin{subfigure}[b]{0.24\textwidth}            
            \includegraphics[width=\textwidth,keepaspectratio=true]{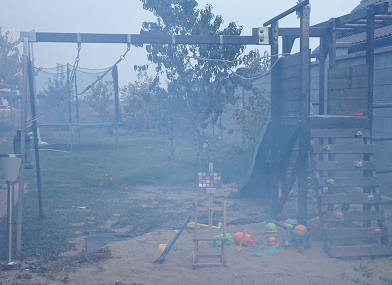}
            \caption*{Input}
    \end{subfigure}
     \hspace{0.1mm}
     \begin{subfigure}[b]{0.24\textwidth}
            \includegraphics[width=\textwidth,keepaspectratio=true]{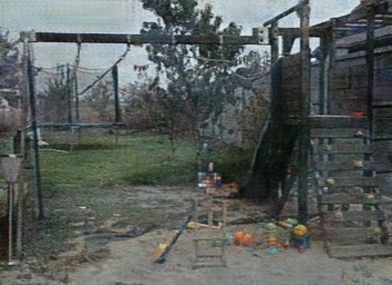}
            \caption*{CycleGAN}
    \end{subfigure}
     \hspace{0.1mm}
     \begin{subfigure}[b]{0.24\textwidth}
            \includegraphics[width=\textwidth,keepaspectratio=true]{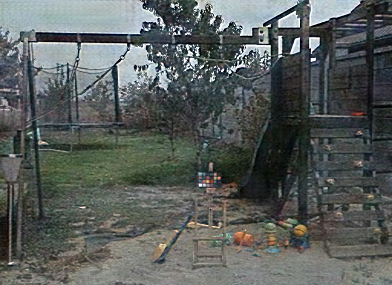}
            \caption*{Ours}
    \end{subfigure}
   \hspace{0.1mm}
    \begin{subfigure}[b]{0.24\textwidth}
            \includegraphics[width=\textwidth,keepaspectratio=true]{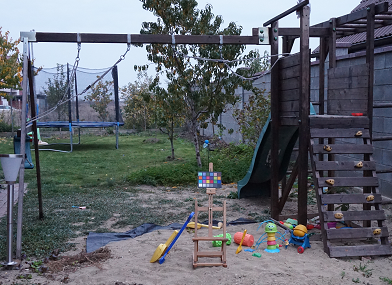}
            \caption*{Ground Truth}
    \end{subfigure}
    \caption{Qualitative results on I-HAZE~\cite{indoorNTIREdataset} \& O-HAZE~\cite{outdoorNTIREdataset} datasets.}
    \label{figure:ntire}
\end{figure*}

\begin{table*}[t!]
\centering
\begin{tabular}{cc|ccccccc}
\hline
\multirow{2}{*}{} & \multirow{2}{*}{} & \multicolumn{2}{c}{\textbf{I-HAZE~\cite{indoorNTIREdataset}}} & \multicolumn{2}{c}{\textbf{O-HAZE~\cite{outdoorNTIREdataset}}} \\ 
        Methods    &  Metrics  & Validation & Test & Validation & Test  \\ 
        \hline \hline
\multirow{1}{*}{None} & PSNR / SSIM & 13.80 / 0.73 & - / - & 13.56/0.59 & - / -    \\  \hline 
\multirow{1}{*}{Best} & PSNR / SSIM & 20.41 / 0.85 & 24.97 / 0.88 & 22.82 / 0.74 & 24.59 / 0.77    \\  \hline
\multirow{1}{*}{CycleGAN} & PSNR / SSIM & 17.80 / 0.75 & - / - & 18.92 / 0.53 & - / -\\ \hline
\multirow{1}{*}{\textbf{Ours}} & PSNR / SSIM & 18.03 / 0.80 & 18.30 / 0.80 & 19.92 / 0.64 & 19.62 / 0.67 \\ 
\hline 
\end{tabular}
\caption{Average PSNR and SSIM results on the NTIRE 2018 challenge on single image dehazing~\cite{ChallengeReportAncuti_2018_CVPR_Workshops} datasets: I-HAZE~\cite{indoorNTIREdataset} \& O-HAZE~\cite{outdoorNTIREdataset}. According to preliminary results, the first row demonstrates the best accuracies of the NTIRE 2018 challenge on single image dehazing~\cite{ChallengeReportAncuti_2018_CVPR_Workshops}. The first row of the results shows the values which are average PSNR and SSIM results calculated directly between the each hazy and its ground truth image.}
\label{table:ntire}
\end{table*}

\textbf{Data Augmentation.} We have employed data augmentation by taking random crops as the pre-processing step before the training phase. This procedure makes our model robust for different scales and textures. Our data augmentation procedure is as in Algorithm~\ref{augmentation}.
\begin{algorithm}[h!]
\caption{Data Augmentation}\label{augmentation}
\begin{algorithmic}[1]
\Procedure{augmenter}{}
\State $\textit{factor} \gets \text{the number of augmented image}$
\State $\text{Image} \text{ } \textit{crops[1..factor]} \gets \textit{[]}$
\State $i \gets 0$

\BState \emph{loop}:
\If {$i = \textit{factor}$}
\State $\text{WRITE(crops); \textbf{break}}$
\EndIf

\State $\textit{x, y} \gets \text{select a random pixel coordinate on image}$
\State $\textit{w, h} \gets \text{select a random width, height}$
\State $\textit{crops}(i) \gets \text{CROP(x,y,w,h)}$
\State $\textit{crops}(i) \gets \text{RESIZE(\textit{crops}(i), [256, 256])}$
\State $i \gets i+1$.
\State \textbf{goto} \emph{loop}.
\EndProcedure
\end{algorithmic}
\end{algorithm}

According to Algorithm \ref{augmentation}, we take random crops from the image by selecting random pixel coordinates and crop sizes. Then, we resize our crops to $256 \times 256$ before feeding them into our network. We run our \emph{augmenter} function for each image in our datasets.

During data augmentation, we have obtained $200$ images per original image in the training set of the I-HAZE~\cite{indoorNTIREdataset} and O-HAZE~\cite{outdoorNTIREdataset} datasets, since this dataset contains too few images.  

\subsection{Results on NYU-Depth Dataset}

We have conducted our experiments on the benchmark NYU-Depth~\cite{NYUdataset} dataset to illustrate the performance of our approach compared to the other \emph{state-of-the-art} approaches. We have employed Cycle-Dehaze by taking hazy images as input, and compared haze-free outputs with theirs ground truths. We have tested our method on all images of NYU-Depth~\cite{NYUdataset} dataset and reported average PSNR and SSIM values in Table \ref{table:d-hazy}. 


Table \ref{table:d-hazy} compares our quantitative results with the other approaches including CycleGAN~\cite{cyclegan}. By outperforming the \emph{state-of-the-art} methods~\cite{dcp2011,CAP2015fast,berman2016non,cai2016dehazenet,reneccv2016singleCNN} according to PSNR metric, our model achieves the second best result. Moreover, Cycle-Dehaze reaches higher PSNR and SSIM values than CycleGAN~\cite{cyclegan}. This demonstrates that adding perceptual-consistency loss on CycleGAN~\cite{cyclegan} improves the architecture further for PSNR and SSIM metrics on NYU-Depth~\cite{NYUdataset} dataset. The results also indicate that Cycle-Dehaze could get nearly similar PSNR results with the approaches profited from parameters of the atmospheric scattering model~\cite{yang2018towards}. 


\subsection{Results on I-HAZE and O-HAZE Datasets}

We have focused on the NTIRE 2018 challenge on single image dehazing~\cite{ChallengeReportAncuti_2018_CVPR_Workshops} datasets: I-HAZE~\cite{indoorNTIREdataset} \& O-HAZE~\cite{outdoorNTIREdataset} during the preparation of this work. We have analyzed effects of Laplacian pyramid and cyclic perceptual-loss, especially on I-HAZE~\cite{indoorNTIREdataset} dataset. The challenge datasets are considerably higher resolution than other image dehazing datasets \eg NYU-Depth~\cite{NYUdataset}. Therefore, the scaling process on images has a larger effect on I-HAZE~\cite{indoorNTIREdataset} and O-HAZE~\cite{outdoorNTIREdataset} datasets according to PSNR and SSIM metrics. Our Laplacian pyramid reduces this deforming effect of the scaling process. We have tested our method on all validation and test images of the challenge datasets provided by organizers of the NTIRE 2018 challenge on single image dehazing~\cite{ChallengeReportAncuti_2018_CVPR_Workshops}. We have trained our final model only on the training set, which is also provided by the organizers. Table \ref{table:ntire} presents average PSNR and SSIM values and Figure \ref{figure:ntire} shows sample qualitative results.\\


 \begin{figure*}[ht!]
\centering
    \begin{subfigure}[b]{0.13\textwidth}            
            \includegraphics[width=\textwidth,keepaspectratio=true]{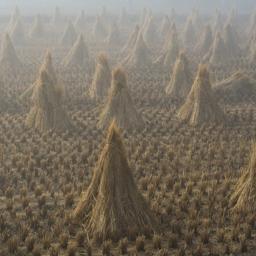}
    \end{subfigure}
    \hspace{0.05mm}
    \begin{subfigure}[b]{0.13\textwidth}            
            \includegraphics[width=\textwidth,keepaspectratio=true]{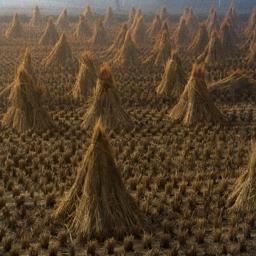}
    \end{subfigure}
    \hspace{0.05mm}
    \begin{subfigure}[b]{0.13\textwidth}            
            \includegraphics[width=\textwidth,keepaspectratio=true]{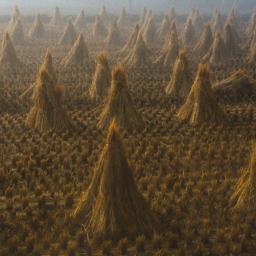}
    \end{subfigure}
    \hspace{0.05mm}
    \begin{subfigure}[b]{0.13\textwidth}            
            \includegraphics[width=\textwidth,keepaspectratio=true]{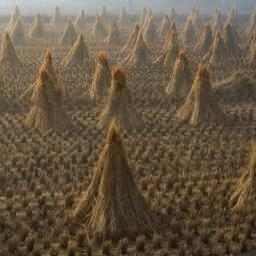}
    \end{subfigure}
    \hspace{0.05mm}
    \begin{subfigure}[b]{0.13\textwidth}            
            \includegraphics[width=\textwidth,keepaspectratio=true]{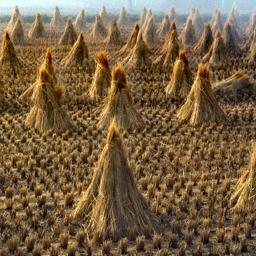}
    \end{subfigure}
    \hspace{0.05mm}
    \begin{subfigure}[b]{0.13\textwidth}            
            \includegraphics[width=\textwidth,keepaspectratio=true]{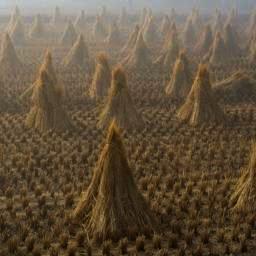}
    \end{subfigure}
    \hspace{0.05mm}
    \begin{subfigure}[b]{0.13\textwidth}            
            \includegraphics[width=\textwidth,keepaspectratio=true]{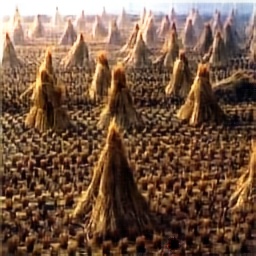}
    \end{subfigure}
     \\
      \vspace{1mm}
    \begin{subfigure}[b]{0.13\textwidth}            
            \includegraphics[width=\textwidth,keepaspectratio=true]{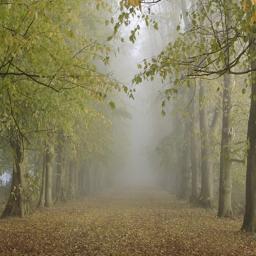}
    \end{subfigure}
    \hspace{0.05mm}
    \begin{subfigure}[b]{0.13\textwidth}            
            \includegraphics[width=\textwidth,keepaspectratio=true]{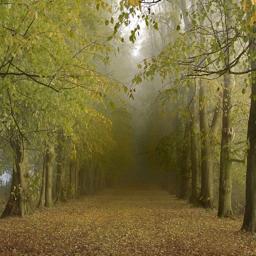}
    \end{subfigure}
    \hspace{0.05mm}
    \begin{subfigure}[b]{0.13\textwidth}            
            \includegraphics[width=\textwidth,keepaspectratio=true]{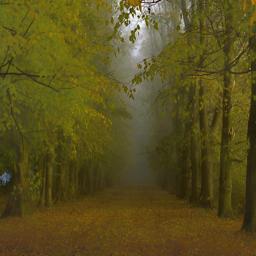}
    \end{subfigure}
    \hspace{0.05mm}
    \begin{subfigure}[b]{0.13\textwidth}            
            \includegraphics[width=\textwidth,keepaspectratio=true]{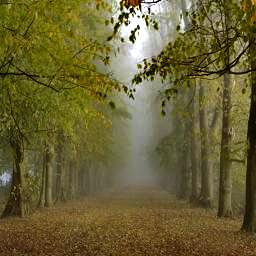}
    \end{subfigure}
    \hspace{0.05mm}
    \begin{subfigure}[b]{0.13\textwidth}            
            \includegraphics[width=\textwidth,keepaspectratio=true]{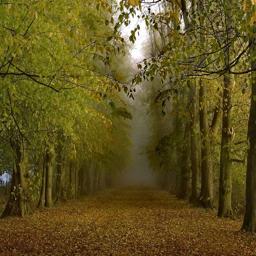}
    \end{subfigure}
    \hspace{0.05mm}
    \begin{subfigure}[b]{0.13\textwidth}            
            \includegraphics[width=\textwidth,keepaspectratio=true]{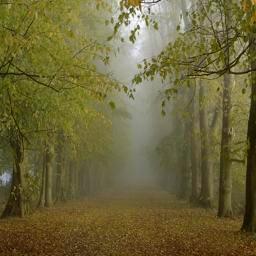}
    \end{subfigure}
    \hspace{0.05mm}
    \begin{subfigure}[b]{0.13\textwidth}            
            \includegraphics[width=\textwidth,keepaspectratio=true]{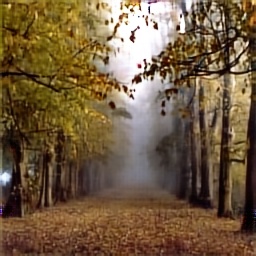}
    \end{subfigure}
    \\
    \vspace{1mm}
    \begin{subfigure}[b]{0.13\textwidth}            
            \includegraphics[width=\textwidth,keepaspectratio=true]{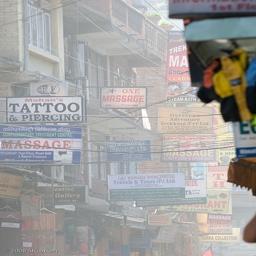}
    \end{subfigure}
    \hspace{0.05mm}
    \begin{subfigure}[b]{0.13\textwidth}            
            \includegraphics[width=\textwidth,keepaspectratio=true]{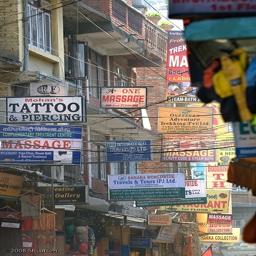}
    \end{subfigure}
    \hspace{0.05mm}
    \begin{subfigure}[b]{0.13\textwidth}            
            \includegraphics[width=\textwidth,keepaspectratio=true]{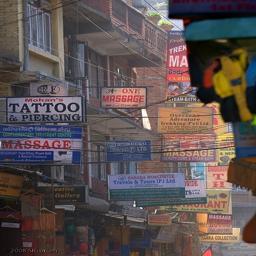}
    \end{subfigure}
    \hspace{0.05mm}
    \begin{subfigure}[b]{0.13\textwidth}            
            \includegraphics[width=\textwidth,keepaspectratio=true]{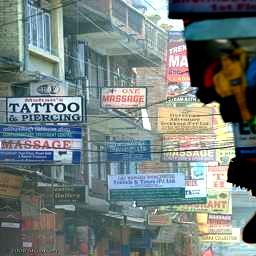}
    \end{subfigure}
    \hspace{0.05mm}
    \begin{subfigure}[b]{0.13\textwidth}            
            \includegraphics[width=\textwidth,keepaspectratio=true]{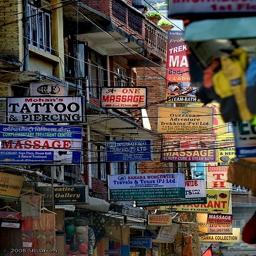}
    \end{subfigure}
    \hspace{0.05mm}
    \begin{subfigure}[b]{0.13\textwidth}            
            \includegraphics[width=\textwidth,keepaspectratio=true]{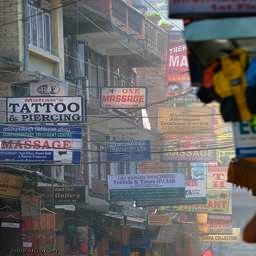}
    \end{subfigure}
    \hspace{0.05mm}
    \begin{subfigure}[b]{0.13\textwidth}            
            \includegraphics[width=\textwidth,keepaspectratio=true]{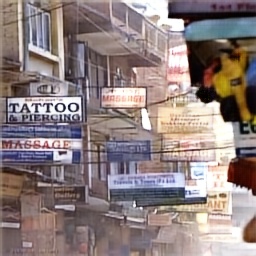}
    \end{subfigure}
    \\
     \vspace{1mm}
    \begin{subfigure}[b]{0.13\textwidth}            
            \includegraphics[width=\textwidth,keepaspectratio=true]{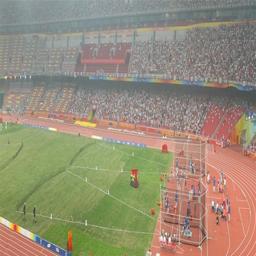}
    \end{subfigure}
    \hspace{0.05mm}
    \begin{subfigure}[b]{0.13\textwidth}            
            \includegraphics[width=\textwidth,keepaspectratio=true]{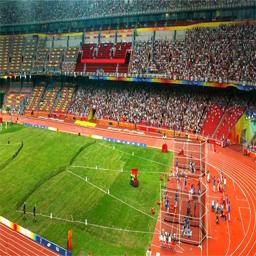}
    \end{subfigure}
    \hspace{0.05mm}
    \begin{subfigure}[b]{0.13\textwidth}            
            \includegraphics[width=\textwidth,keepaspectratio=true]{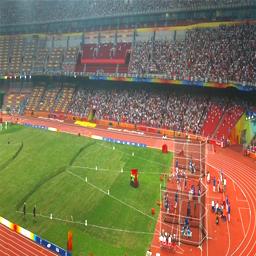}
    \end{subfigure}
    \hspace{0.05mm}
    \begin{subfigure}[b]{0.13\textwidth}            
            \includegraphics[width=\textwidth,keepaspectratio=true]{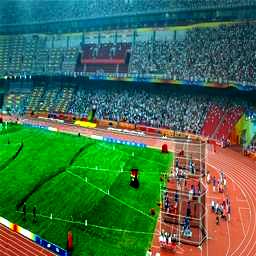}
    \end{subfigure}
    \hspace{0.05mm}
    \begin{subfigure}[b]{0.13\textwidth}            
            \includegraphics[width=\textwidth,keepaspectratio=true]{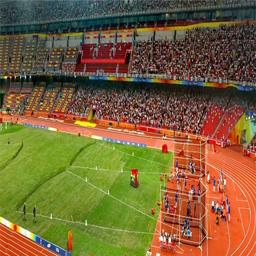}
    \end{subfigure}
    \hspace{0.05mm}
    \begin{subfigure}[b]{0.13\textwidth}            
            \includegraphics[width=\textwidth,keepaspectratio=true]{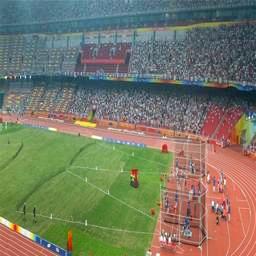}
    \end{subfigure}
    \hspace{0.05mm}
    \begin{subfigure}[b]{0.13\textwidth}            
            \includegraphics[width=\textwidth,keepaspectratio=true]{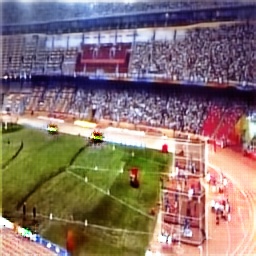}
    \end{subfigure}
    \\
     \vspace{1mm}
    \begin{subfigure}[b]{0.13\textwidth}            
            \includegraphics[width=\textwidth,keepaspectratio=true]{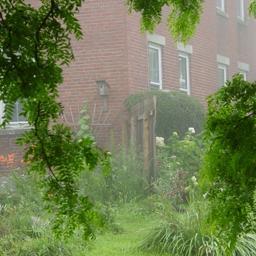}
             \caption*{Input}
    \end{subfigure}
    \hspace{0.05mm}
    \begin{subfigure}[b]{0.13\textwidth}            
            \includegraphics[width=\textwidth,keepaspectratio=true]{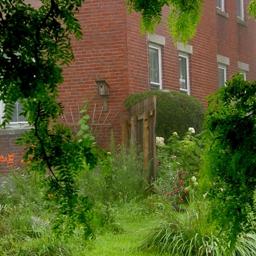}
             \caption*{He \etal~\cite{dcp2011}}
    \end{subfigure}
    \hspace{0.05mm}
    \begin{subfigure}[b]{0.13\textwidth}            
            \includegraphics[width=\textwidth,keepaspectratio=true]{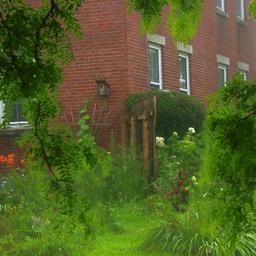}
             \caption*{Zhu \etal~\cite{CAP2015fast}}
    \end{subfigure}
    \hspace{0.05mm}
    \begin{subfigure}[b]{0.13\textwidth}            
            \includegraphics[width=\textwidth,keepaspectratio=true]{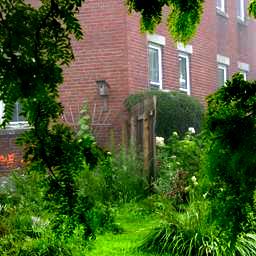}
             \caption*{Ren \etal~\cite{reneccv2016singleCNN}}
    \end{subfigure}
    \hspace{0.05mm}
    \begin{subfigure}[b]{0.13\textwidth}            
            \includegraphics[width=\textwidth,keepaspectratio=true]{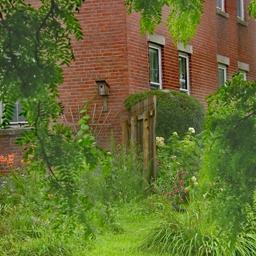}
             \caption*{Berman \etal~\cite{berman2016non}}
    \end{subfigure}
    \hspace{0.05mm}
    \begin{subfigure}[b]{0.13\textwidth}            
            \includegraphics[width=\textwidth,keepaspectratio=true]{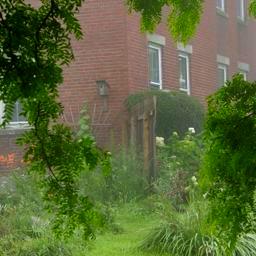}
             \caption*{Cai \etal~\cite{cai2016dehazenet}}
    \end{subfigure}
    \hspace{0.05mm}
    \begin{subfigure}[b]{0.13\textwidth}            
            \includegraphics[width=\textwidth,keepaspectratio=true]{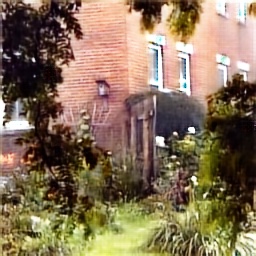}
             \caption*{Cycle-Dehaze}
    \end{subfigure}
    \caption{Qualitative results on natural hazy images by comparing with state-of-the-art-results.}
    \label{figure:natural}
\end{figure*}
 

\textbf{Quantitative Results.} According to Table \ref{table:ntire}, our proposed method gave better PSNR and SSIM values than CycleGAN~\cite{cyclegan} for each track of the challenge. This shows that additional cyclic perceptual-consistency loss and Laplacian pyramid increase the performance of original CycleGAN~\cite{cyclegan} architecture. Moreover, PSNR and SSIM differences between the I-HAZE~\cite{indoorNTIREdataset} and O-HAZE~\cite{outdoorNTIREdataset} datasets presents that outdoor scenes suffer from SSIM values because of the long shot of the captured images. On the other hand, they have higher PSNR values since the produced fog spreads the atmosphere and the captured images seem less hazed than the indoor scenes.

\textbf{Qualitative Results.} Figure \ref{figure:ntire} shows the qualitative difference between CycleGAN~\cite{cyclegan} and Cycle-Dehaze. From the qualitative results, it can be clearly seen that dehazed images by Cycle-Dehaze has less noise and sharper edges for both indoor and outdoor scenes, where cyclic perceptual-consistency loss reduces the noise of the dehazed images and Laplacian pyramid leads sharper edges. Since outdoor scenes include more textural repetitions than indoor scenes, recovering textures of O-HAZE~\cite{outdoorNTIREdataset} is harder than I-HAZE~\cite{indoorNTIREdataset}. Therefore, our sharpness on edges reduces in outdoor conditions.

\subsection{Results on Cross-Dataset Image Dehazing}
\begin{figure}[t!]
 \centering
     \begin{subfigure}[b]{0.23\textwidth}
             \includegraphics[width=\textwidth,height=24mm]{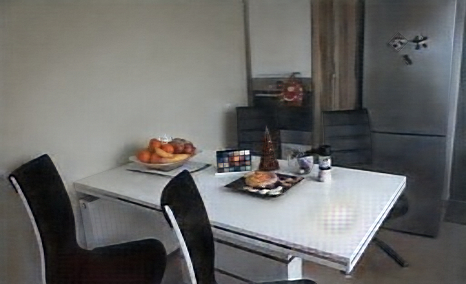}
     \end{subfigure}
     \begin{subfigure}[b]{0.23\textwidth}
             \includegraphics[width=\textwidth,height=24mm]{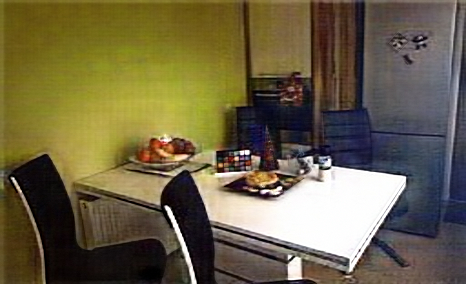}
     \end{subfigure}
     \\
     \vspace{0.5mm}
     \begin{subfigure}[b]{0.23\textwidth}            
             \includegraphics[width=\textwidth,height=24mm]{indoor/27_our}
                  \caption*{Single-dataset}
     \end{subfigure}
      \begin{subfigure}[b]{0.23\textwidth}
             \includegraphics[width=\textwidth,height=24mm]{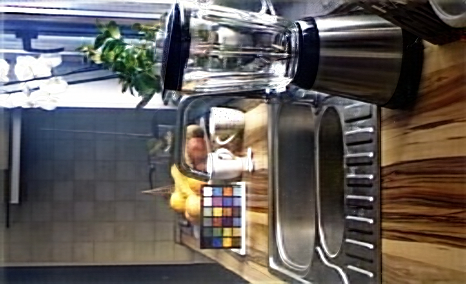}
             \caption*{Cross-dataset}
     \end{subfigure}
     \caption{Comparative qualitative results between single and cross dataset experiments via Cycle-Dehaze.}
     \label{figure:cross}
 \end{figure}

CNNs mostly tend to overfit on a specific dataset rather than learning the targeted task. To the best of our know-ledge, fine-tuning the trained model on a targeted dataset is the most popular solution of overfitting. On the other hand, we have analyzed our method with two distinct experiments in cross-dataset setups, in which entirely different datasets have been used for the training and testing phases. Firstly, we have tested Cycle-Dehaze on some popular natural images used by image dehazing community by scaling them to $256 \times 256$. Figure \ref{figure:natural} provides the qualitative results obtained on them. Secondly, we have tested the final model trained on NYU-Depth~\cite{NYUdataset} dataset on I-HAZE~\cite{indoorNTIREdataset} dataset and vice versa, since both of the datasets have been created under indoor conditions. Table \ref{table:cross} presents the accuracies of cross-dataset testing and Figure \ref{figure:cross} shows the visual difference of cross-dataset testing between the datasets captured at indoor scenes: NYU-Depth~\cite{NYUdataset} and I-HAZE~\cite{indoorNTIREdataset} datasets.

\begin{table}[h!]
  \begin{center}
  \begin{tabular}{cc|cc}
   \hline
   Training set & Test set & PSNR & SSIM  \\
  \hline\hline
  NYU-Depth~\cite{NYUdataset} & NYU-Depth~\cite{NYUdataset} & 15.41 & 0.66 \\
  I-HAZE~\cite{indoorNTIREdataset} & NYU-Depth~\cite{NYUdataset} & 13.12 & 0.59 \\
  I-HAZE~\cite{indoorNTIREdataset} & I-HAZE~\cite{indoorNTIREdataset} & 18.03 & 0.80 \\
  NYU-Depth~\cite{NYUdataset} & I-HAZE~\cite{indoorNTIREdataset} & 14.76 & 0.73\\
  \hline
   \end{tabular}
   \end{center}
    \vspace{-3mm}
   \caption{Cross-dataset quantitative results of Cycle-Dehaze architecture on the datasets captured at indoor scenes.} 
  \label{table:cross}
 \end{table}
\textbf{Quantitative Results.} According to Table \ref{table:cross}, Cycle-Dehaze obtains considerably high PSNR and SSIM values on cross-dataset testing. As a matter of fact, the results both on NYU-Depth~\cite{NYUdataset} and I-HAZE~\cite{indoorNTIREdataset} are as fine as original CycleGAN~\cite{cyclegan} architecture on regular single dataset testing. This shows that Cycle-Dehaze mostly learns the dehazing task rather than overfitting on a dataset. Due to the cyclic mechanism of Cycle-Dehaze, our method focus on adding a haze on images beside cleaning a haze. Therefore, Cycle-Dehaze learns what is haze regardless of the image dehazing problem. On the other side, the methods only addressed the dehazing process tend to focus on color enhancement on the specific dataset. From the result of cross-dataset experiments, Cycle-Dehaze can be considered as a practical solution on real-world conditions for single image dehazing.\\ 

\textbf{Qualitative Results.} Figure \ref{figure:natural} shows the comparative qualitative results of Cycle-Dehaze on natural hazy images with respect to state-of-the-art image dehazing methods \cite{dcp2011,CAP2015fast,berman2016non,reneccv2016singleCNN,cai2016dehazenet}. According to qualitative results, the performance of Cycle-Dehaze is perceptually satisfying on natural images, especially when the color tones of neighboring pixels are very close to each other. Specifically, Cycle-Dehaze preserves the natural color toning of hazy image after dehazing process. Consequently, Cycle-Dehaze keeps the shadows and depth on the image more perceptible.

Figure \ref{figure:cross} includes the images, which are dehazed by regular Cycle-Dehaze and by the cross-dataset version of it. According to qualitative results, both methods can clear the haze from the input images. On the other hand, the color recovery on the single dataset is better than on cross-dataset which leads lower PSNR results on cross-dataset scenario. Since the haze thickens at some parts of the images, our model can not estimate the actual ground truth color if it is trained on another dataset.

\section{Conclusion} \label{conc}

We proposed a single image dehazing network, named as Cycle-Dehaze, which directly generates haze-free images from hazy input images without estimating parameters of the atmospheric scattering model. Besides, our network provides a training process of hazy and ground truth images in an unpaired manner. In order to retain the high visual quality of haze-free images, we improved cycle-consistency loss of CycleGAN architecture by combining it with the perceptual loss. Cycle-Dehaze takes low-resolution images as input, so it requires downscaling of its inputs as a pre-processing step. For reducing distortion on images while resizing, we utilized Laplacian pyramid to upscale low-resolution images instead of using directly bicubic upscaling. 
The experimental results show that our method produces visually better images and achieves higher PSNR and SSIM values than CycleGAN architecture. Moreover, we performed additional experiments on the cross-dataset scenario to demonstrate generalizability of our model for different domains. \\

\noindent\textbf{Acknowledgements.} We would like to thank our colleagues from SiMiT Lab at ITU and LTS5 at EPFL, especially Christophe Ren\`e Joseph Ecabert and Saleh Bagher Salimi, for their valuable comments. Travel grant for this research is provided by Yap{\i} Kredi Technology. 

{\small
\bibliographystyle{ieee}
\bibliography{egbib}
}
\end{document}